\documentclass[conference]{IEEEtran}
\IEEEoverridecommandlockouts
\usepackage{cite}
\usepackage{amsmath,amssymb,amsfonts}
\usepackage{algorithmic}
\usepackage{graphicx}
\usepackage{textcomp}
\usepackage{xcolor}
\usepackage{multirow}
\usepackage{array}
\usepackage{balance}
\def\BibTeX{{\rm B\kern-.05em{\sc i\kern-.025em b}\kern-.08em
		T\kern-.1667em\lower.7ex\hbox{E}\kern-.125emX}}
\makeatletter
\newcommand{\linebreakand}{%
\end{@IEEEauthorhalign}
\hfill\mbox{}\par
\mbox{}\hfill\begin{@IEEEauthorhalign}
}
\makeatother
\begin{document}
\bstctlcite{IEEEexample:BSTcontrol}

\title{MCS-HMS: A Multi-Cluster Selection Strategy for the Human Mental Search Algorithm\thanks{This research is funded within the SFEDU development program (PRIORITY 2030).}
}

\author{
\IEEEauthorblockN{Ehsan Bojnordi}
\IEEEauthorblockA{\textit{Information Technology Department} \\
\textit{Iranian National Tax Administration}\\
Bojnord, Iran} 
\and
\IEEEauthorblockN{Seyed Jalaleddin Mousavirad}
\IEEEauthorblockA{\textit{Computer Engineering Department}\\
\textit{Hakim Sabzevari University}\\
Sabzevar, Iran}
\and
\IEEEauthorblockN{Gerald Schaefer }
\IEEEauthorblockA{\textit{Department of Computer Science} \\
\textit{Loughborough University}\\
Loughborough, U.K.}
\linebreakand
\IEEEauthorblockN{Iakov Korovin}
\IEEEauthorblockA{\textit{Southern Federal University}\\
Taganrog, Russia}\\
}
\maketitle

\begin{abstract}
Population-based metaheuristic algorithms have received significant attention in global optimisation. Human Mental Search (HMS) is a relatively recent population-based metaheuristic that has been shown to work well in comparison to other algorithms. However, HMS is time-consuming and suffers from relatively poor exploration. Having clustered the candidate solutions, HMS selects a winner cluster with the best mean objective function. This is not necessarily the best criterion to choose the winner group and limits the exploration ability of the algorithm.

In this paper, we propose an improvement to the HMS algorithm in which the best bids from multiple clusters are used to benefit from enhanced exploration. We also use a one-step $k$-means algorithm in the clustering phase to improve the speed of the algorithm. Our experimental results show that MCS-HMS outperforms HMS as well as other population-based metaheuristic algorithms.
\end{abstract}

\begin{IEEEkeywords}
Global optimisation, metaheuristic algorithms, Human Mental Search, clustering
\end{IEEEkeywords}

\section{Introduction}
Many optimisation problems can be formulated as searching for the best element or solution from all feasible solutions with respect to one or several criteria. While conventional optimisation algorithms like gradient-based methods are rooted in a solid mathematical formulation, they typically fail to provide satisfactory results and often stagnate in local optima~\cite{1-DE_Shift_Mechanism},\cite{2-8845065},\cite{3-10.1504/ijbic.2019.103961}.

Population-based metaheuristic algorithms are able to address this issue~\cite{4-hussain2019metaheuristic} and can be divided into three general categories, namely evolutionary, swarm-based, and physics-based approaches. Evolutionary algorithms, such as the genetic algorithm (GA)~\cite{5-GA_Main_Ref}, are inspired by biological evolutionary processes. Swarm-based algorithms, e.g.\ particle swarm optimization (PSO)~\cite{6-PSO_Main_Paper} and artificial bee colony (ABC)~\cite{7-ABC_Main_Paper}, imitate the social behaviour of animals, while physics-based algorithms, including the gravitational search algorithm (GSA)~\cite{8-rashedi2009gsa} and the mine blast algorithm (MBA)~\cite{9-SADOLLAH20132592}, are based on the laws of physics.

Human Mental Search (HMS)~\cite{11-HMS_Main_Paper} is a relatively recent population-based metaheuristic algorithm that is based on the concept of exploring the search space of online auctions. Here, mental search, a main part of the algorithm, employs Levy flight to explore the vicinity of candidate solutions. HMS has been shown to solve effectively a wide range of optimisation problems, including unimodal, multi-modal, high-dimensional, rotated, shifted, and complex functions~\cite{11-HMS_Main_Paper}, as well as various machine vision applications including multi-level thresholding~\cite{13-mousavirad2020human,13-1-ESMAEILI2021115106}, colour quantisation~\cite{13-2-10.1007/978-3-030-53956-6_12,13-3-9283370}, image segmentation~\cite{13-4-MOUSAVIRAD2020106604}, and image clustering~\cite{12-HMS_Image_Clustering}. 

HMS algorithm has three main operators, mental search to explore the vicinity of candidate solutions based on a Levy flight distribution, grouping to cluster the current population to find a promising region, and movement to move candidate solutions towards the promising region. Several improvements to HMS have been recently introduced, including leveraging a random clustering strategy~\cite{14-1-8914636}, and grouping in both search and objective space~\cite{14-2-8914636}.

In the grouping phase of HMS, a clustering algorithm is employed and the cluster with the best mean objective function value is selected as the winner cluster. However, this approach has a tendency of getting stuck in a local optimum. To address this issue, in this paper, we propose a novel HMS algorithm, MCS-HMS, that is based on multi-cluster selection. Here, the best candidate solutions in each cluster have a chance of being selected. We further employ a more efficient one-step $k$-means algorithm for clustering. Extensive experiments on various benchmark functions with different characteristics show that MCS-HMS outperforms HMS as well as other population-based metaheuristic algorithms.

The remainder of the paper is organised as follows. Section~\ref{sec:HMS} describes the standard HMS algorithm. Our proposed MCS-HMS is then introduced in Section~\ref{sec:MCS-HMS}, while experimental results are given in Section~\ref{sec:Exp}. Finally, Section~\ref{sec:Concl} concludes the paper.

\section{Human Mental Search}
\label{sec:HMS}
HMS is a metaheuristic algorithm inspired by the way the human mind searches. Each candidate solution, called a bid in HMS, is initially randomly generated. HMS then proceeds, iteratively, based on three main operators, mental search, grouping, and movement. 

During mental search, a number of new bids are obtained as 
\begin{equation}
x^i + S ,
\end{equation}
where $x^i$ is a current bid, and
\begin{equation}
S=(2-NFE(\frac{2}{{NFE}_{\max}}))0.01\frac{u}{v^{\frac{1}{\beta}}}(x^i-x^*) ,
\end{equation}
where $NFE$ represents the number of objective function evaluations thus far, ${NFE}_{\max}$ is the maximum number of function evaluations, $\beta$ is a random integer number, and $x^*$ is the best bid found so far. $u$ and $v$ are two random numbers with normal distributions as
\begin{equation}
u{\approx}N(0,\sigma_u^2 ),~~~~~~v{\approx}N(0,\sigma_v^2 ), 
\end{equation}
and
\begin{equation}
{\sigma}_{u}=(\frac{\Gamma(1+{\beta})sin(\frac{\pi\beta}{2})}{\Gamma(\frac{1+\beta}{2})\beta2^{\frac{\beta-1}{2}}})^{\frac{1}{\beta}},~~~~~~{\sigma}_{v}=1,
\end{equation}
where $\Gamma$ is a standard gamma function.

The grouping operator uses the $k$-means algorithm to cluster similar bids in the population. Then, for each cluster the mean objective function value is calculated and the cluster with the best value is selected as the winner cluster to represent a promising area in search space. 

Bids in the other clusters then towards the identified promising area by
\begin{equation}
x^i_n(t+1) = x^i_n(t) + C(r W_n(t) - X^i_n(t)),
\end{equation}
where $W$ is the best solution in the promising area, $C$ is a constant, $r$ is a random number in $[0;1]$, $t$ indicates the current iteration, and subscript $n$ indicates the $n$-th element of a bid.

\section{Proposed MCS-HMS Algorithm}
\label{sec:MCS-HMS}
One of the drawbacks of the standard HMS algorithm is that selecting the cluster with the best mean objective functions does not necessarily give the best choice for the promising area because the cluster might be placed far away from the global optimum. Therefore, choosing the best cluster and its best bid may limit diversification and may cause the next generations to be misled to a less successful area. Therefore, in this paper, we introduce a novel modification of HMS to tackle this issue and find a better promising area.

\begin{figure}[h!]
\centering
\includegraphics[width=.9\columnwidth]{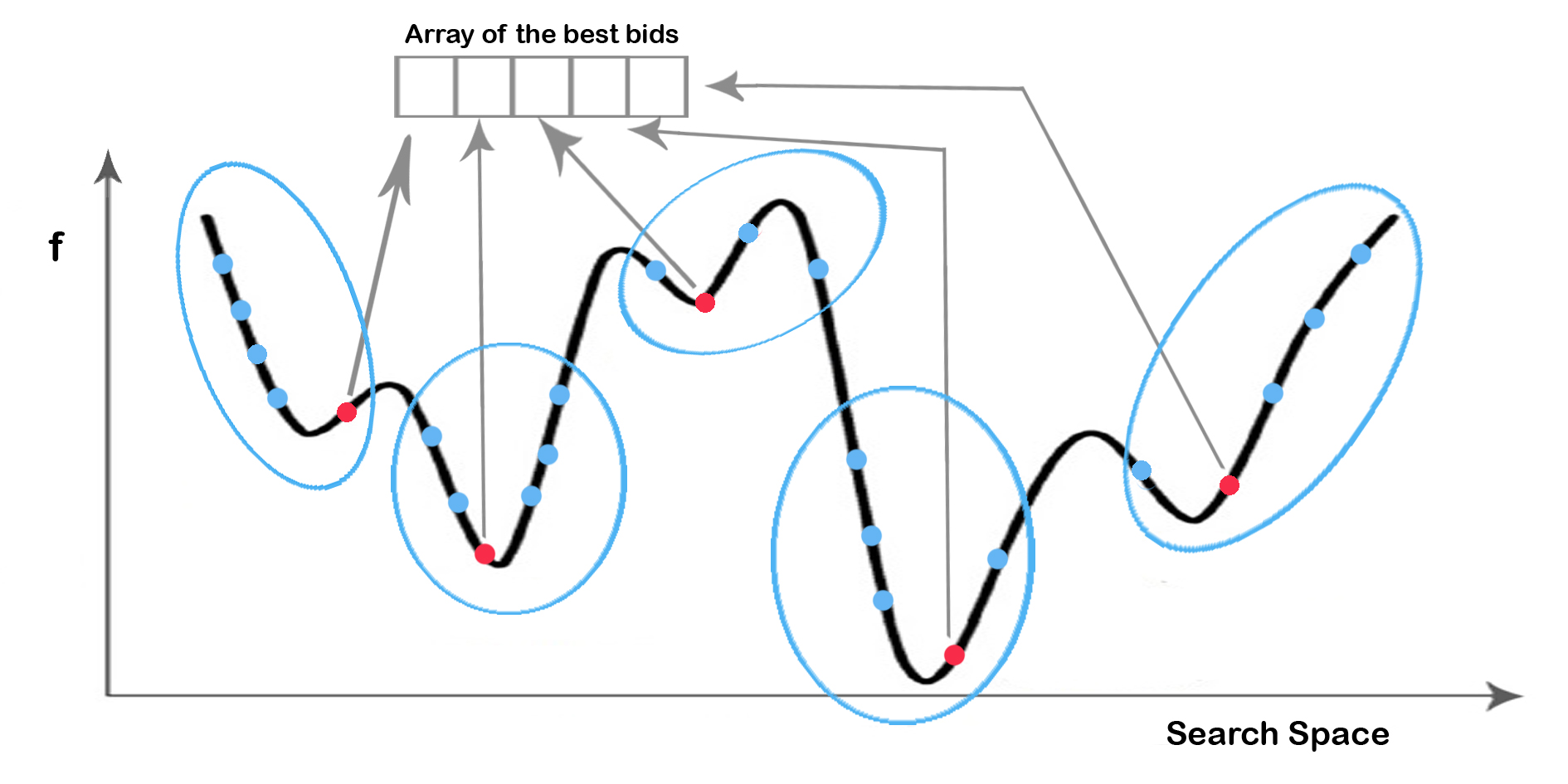}
\caption{Multi-cluster selection process after grouping.}
\label{fig1}
\end{figure}

Our idea is to select a few promising bids in a memory $M$ and use these. Specifically, instead of selecting the best bid in the cluster with the best mean objective function, we consider the best bid of each cluster as illustrated in Fig~\ref{fig1}. The memory keeps only the bids with the best objective function value of each cluster, and thus the length of $M$ equals the number of clusters.

In the next step, we randomly select a bid from $M$ as the target for the movement operator and thus the corresponding cluster as the promising area in search space. In contrast to standard HMS, this mechanism leads to improved exploration since the target is selected from the whole search space, and can also prevent early convergence.

In addition, we address the relative inefficiency of HMS due to its application of $k$-means in the grouping process. Inspired by \cite{14-9185525}, In MCS-HMS we replace $k$-means with a one-step $k$-means algorithm. That is only a single iteration of $k$-means is conducted, leading to quicker execution.

\section{Experimental Results}
\label{sec:Exp}
For evaluation, we use the 30 CEC 2017 benchmark functions~\cite{15-CEC2017} which include unimodal (F1 to F3), multi-modal (F4 to F10), hybrid multi-modal (F11 to F20) and composite (F21 to F30) test functions. Besides comparing MCS-HMS with standard HMS, we also benchmark it against a number of other population-based metaheuristic algorithms, namely particle swarm optimisation (PSO)~\cite{6-PSO_Main_Paper}, the covariance matrix adaption-evolution strategy (CMA-ES)~\cite{18-hansen2001completely}, artificial bee colony (ABC)~\cite{7-ABC_Main_Paper}, the grey wolf optimiser (GWO)~\cite{10-GWO_Main_Paper}, moth flame optimisation (MFO)~\cite{16-MFO_Main_Paper}, and the whale optimisation algorithm (WOA)~\cite{17-WOA_Main_Paper}.

As dimensionality $D$ of the search space we use 30, 50 and 100. The employed parameters of the various algorithms are given in Table~\ref{tab1}.

\begin{table}[b]
\centering
\caption{Parameter settings of experiments.}
\label{tab1}
\begin{tabular}{llc}
\hline
algorithm & parameter & values \\
\hline
PSO & inertia constant & 1 to 0 \\
& cognitive constant & 2 \\
& social constant & 2 \\
\hline
CMA-ES & $\lambda$ & 50 \\ 
\hline
ABC & limit & $n_e$  × $D$ \\
\hline
GWO & no parameters \\ 
\hline
MFO & no parameters \\ 
\hline
WOA & $b$ & 1 \\
\hline
HMS & number of clusters & 5 \\
& minimum mental processes & 2 \\
& maximum mental processes & 5 \\
& $C$ & 1 \\ 
\hline
MCS-HMS & same settings as HMS \\
\hline
\end{tabular}
\end{table}

\begin{table*}[p]
\centering
\caption{Experimental results for $D=30$.}
\label{tab5}
\begin{tabular}{c|cccccccc}
\hline
& PSO & CMA-ES & ABC & GWO & MFO & WOA & HMS & MCS-HMS \\ 
\hline 
F1 & 7.16E+07 & 2.11E+10 & 2.91E+07 & 1.75E+09 & 8.80E+09 & 8.43E+07 & 4.55E+06 & 1.37E+04 \\
F2 & 8.13E+11 & 3.33E+42 & 3.93E+41 & 4.09E+28 & 1.16E+39 & 1.70E+32 & 4.19E+24 & 6.08E+20 \\
F3 & 3.24E+02 & 2.11E+05 & 3.68E+05 & 3.94E+04 & 1.08E+05 & 1.99E+05 & 1.09E+04 & 1.33E+04 \\
F4 & 7.54E+01 & 3.62E+03 & 1.19E+02 & 1.86E+02 & 6.64E+02 & 2.02E+02 & 1.21E+02 & 9.96E+01 \\
F5 & 2.12E+02 & 3.31E+02 & 2.39E+02 & 9.59E+01 & 1.98E+02 & 2.78E+02 & 1.19E+02 & 9.64E+01 \\
F6 & 5.21E+01 & 6.30E+01 & 5.20E+00 & 7.58E+00 & 3.04E+01 & 6.82E+01 & 1.01E+01 & 3.52E+00 \\
F7 & 2.47E+02 & 1.81E+02 & 2.78E+02 & 1.53E+02 & 3.83E+02 & 5.06E+02 & 1.46E+02 & 1.30E+02 \\
F8 & 1.57E+02 & 2.69E+02 & 2.47E+02 & 8.14E+01 & 1.90E+02 & 2.31E+02 & 1.07E+02 & 9.82E+01 \\
F9 & 4.64E+03 & 1.66E+03 & 3.08E+03 & 9.48E+02 & 5.76E+03 & 7.99E+03 & 1.39E+03 & 7.98E+02 \\
F10 & 4.83E+03 & 7.03E+03 & 8.17E+03 & 3.22E+03 & 4.63E+03 & 5.46E+03 & 3.71E+03 & 3.50E+03 \\
F11 & 1.53E+02 & 1.67E+04 & 9.40E+03 & 8.31E+02 & 3.64E+03 & 1.98E+03 & 2.28E+02 & 1.42E+02 \\
F12 & 1.45E+07 & 4.32E+09 & 4.97E+08 & 6.48E+07 & 2.23E+08 & 8.52E+07 & 7.50E+06 & 2.86E+06 \\
F13 & 1.87E+06 & 3.89E+09 & 2.57E+06 & 1.23E+07 & 4.86E+07 & 2.07E+05 & 4.54E+04 & 3.23E+04 \\
F14 & 1.41E+04 & 5.45E+06 & 2.98E+05 & 2.57E+05 & 1.30E+05 & 2.39E+06 & 4.27E+04 & 3.91E+04 \\
F15 & 1.56E+05 & 4.90E+08 & 1.40E+06 & 1.58E+06 & 4.53E+04 & 1.26E+05 & 7.90E+03 & 5.82E+03 \\
F16 & 1.26E+03 & 3.25E+03 & 2.34E+03 & 9.59E+02 & 1.57E+03 & 2.06E+03 & 9.13E+02 & 1.08E+03 \\
F17 & 5.43E+02 & 1.91E+03 & 1.14E+03 & 2.66E+02 & 7.10E+02 & 9.24E+02 & 4.28E+02 & 3.69E+02 \\
F18 & 2.17E+05 & 3.35E+07 & 1.46E+07 & 1.13E+06 & 3.06E+06 & 4.17E+06 & 5.33E+05 & 4.73E+05 \\
F19 & 5.96E+05 & 4.42E+08 & 8.37E+04 & 8.96E+05 & 9.48E+06 & 4.38E+06 & 1.35E+04 & 1.44E+04 \\
F20 & 5.96E+02 & 8.25E+02 & 9.94E+02 & 3.85E+02 & 6.57E+02 & 8.42E+02 & 3.28E+02 & 3.38E+02 \\
F21 & 4.05E+02 & 5.51E+02 & 4.46E+02 & 2.84E+02 & 3.97E+02 & 4.79E+02 & 3.20E+02 & 3.04E+02 \\
F22 & 3.13E+03 & 7.49E+03 & 8.25E+03 & 2.07E+03 & 4.10E+03 & 5.14E+03 & 3.73E+03 & 2.41E+03 \\
F23 & 8.44E+02 & 7.46E+02 & 6.07E+02 & 4.54E+02 & 5.18E+02 & 7.37E+02 & 5.04E+02 & 4.76E+02 \\
F24 & 8.17E+02 & 7.71E+02 & 6.83E+02 & 5.12E+02 & 5.78E+02 & 8.13E+02 & 5.85E+02 & 5.74E+02 \\
F25 & 3.97E+02 & 1.40E+03 & 4.20E+02 & 4.74E+02 & 8.20E+02 & 4.96E+02 & 3.98E+02 & 3.89E+02 \\
F26 & 2.16E+03 & 5.62E+03 & 3.15E+03 & 2.07E+03 & 3.06E+03 & 5.21E+03 & 2.49E+03 & 2.23E+03 \\
F27 & 5.19E+02 & 6.99E+02 & 5.00E+02 & 5.46E+02 & 5.46E+02 & 7.24E+02 & 5.19E+02 & 5.13E+02 \\
F28 & 4.44E+02 & 3.80E+03 & 5.00E+02 & 5.82E+02 & 1.36E+03 & 5.73E+02 & 5.15E+02 & 4.84E+02 \\
F29 & 1.40E+03 & 2.83E+03 & 2.02E+03 & 9.50E+02 & 1.19E+03 & 2.17E+03 & 9.06E+02 & 8.39E+02 \\
F30 & 1.88E+06 & 4.88E+08 & 4.99E+05 & 6.71E+06 & 1.08E+06 & 2.04E+07 & 3.38E+04 & 1.41E+04 \\
\hline
\end{tabular}
\end{table*}

\begin{table*}[p]
\centering
\caption{Experimental results for $D=50$.}
\label{tab6}
\begin{tabular}{c|cccccccc}
\hline
& PSO & CMA-ES & ABC & GWO & MFO & WOA & HMS & MCS-HMS \\ 
\hline 
F1 & 2.57E+08 & 4.64E+10 & 2.81E+09 & 5.59E+09 & 3.75E+10 & 4.96E+08 & 2.99E+08 & 5.03E+05 \\
F2 & 2.66E+25 & 5.07E+79 & 5.48E+80 & 7.96E+54 & 6.94E+76 & 1.25E+70 & 1.42E+53 & 1.22E+45 \\
F3 & 9.48E+03 & 3.75E+05 & 7.71E+05 & 9.42E+04 & 1.67E+05 & 1.85E+05 & 3.48E+04 & 3.71E+04 \\
F4 & 1.62E+02 & 7.61E+03 & 3.24E+03 & 7.14E+02 & 4.06E+03 & 5.30E+02 & 3.04E+02 & 2.20E+02 \\
F5 & 3.94E+02 & 3.08E+02 & 5.49E+02 & 2.07E+02 & 4.62E+02 & 4.70E+02 & 2.38E+02 & 2.27E+02 \\
F6 & 6.95E+01 & 6.80E+01 & 4.93E+01 & 1.76E+01 & 4.78E+01 & 8.30E+01 & 1.87E+01 & 1.16E+01 \\
F7 & 5.01E+02 & 2.20E+02 & 6.50E+02 & 3.59E+02 & 1.13E+03 & 1.06E+03 & 3.00E+02 & 3.31E+02 \\
F8 & 4.15E+02 & 5.17E+02 & 5.51E+02 & 2.19E+02 & 4.79E+02 & 4.49E+02 & 2.54E+02 & 2.38E+02 \\
F9 & 2.47E+04 & 1.35E+04 & 4.27E+04 & 8.16E+03 & 1.71E+04 & 2.76E+04 & 6.46E+03 & 6.91E+03 \\
F10 & 8.82E+03 & 1.32E+04 & 1.50E+04 & 6.50E+03 & 7.47E+03 & 1.01E+04 & 7.53E+03 & 6.73E+03 \\
F11 & 3.59E+02 & 6.45E+04 & 5.59E+04 & 2.97E+03 & 1.28E+04 & 1.17E+03 & 8.25E+02 & 5.58E+02 \\
F12 & 1.07E+08 & 2.24E+10 & 1.05E+10 & 5.44E+08 & 4.63E+09 & 5.79E+08 & 1.07E+08 & 1.94E+07 \\
F13 & 1.47E+07 & 1.29E+10 & 6.17E+07 & 1.05E+08 & 8.80E+08 & 5.55E+06 & 1.17E+05 & 2.51E+04 \\
F14 & 1.17E+05 & 2.17E+07 & 4.29E+06 & 9.45E+05 & 1.08E+06 & 1.97E+06 & 2.85E+05 & 1.68E+05 \\
F15 & 3.05E+06 & 2.51E+09 & 1.28E+07 & 1.56E+07 & 8.65E+06 & 1.08E+06 & 3.24E+04 & 1.43E+04 \\
F16 & 2.15E+03 & 5.37E+03 & 5.11E+03 & 1.46E+03 & 2.73E+03 & 3.81E+03 & 2.10E+03 & 2.05E+03 \\
F17 & 1.68E+03 & 1.05E+03 & 3.12E+03 & 1.06E+03 & 2.30E+03 & 2.55E+03 & 1.51E+03 & 1.39E+03 \\
F18 & 1.50E+06 & 1.27E+08 & 6.83E+07 & 5.10E+06 & 2.96E+06 & 1.41E+07 & 1.78E+06 & 1.29E+06 \\
F19 & 2.78E+06 & 1.10E+09 & 2.55E+04 & 3.16E+06 & 5.93E+07 & 5.13E+06 & 3.12E+04 & 1.44E+04 \\
F20 & 1.30E+03 & 1.62E+03 & 2.50E+03 & 8.97E+02 & 1.40E+03 & 1.70E+03 & 1.10E+03 & 1.08E+03 \\
F21 & 6.61E+02 & 7.99E+02 & 7.52E+02 & 3.94E+02 & 6.41E+02 & 8.43E+02 & 4.91E+02 & 4.71E+02 \\
F22 & 9.09E+03 & 1.43E+04 & 1.51E+04 & 6.78E+03 & 8.28E+03 & 1.01E+04 & 7.85E+03 & 6.99E+03 \\
F23 & 1.55E+03 & 1.14E+03 & 9.78E+02 & 6.64E+02 & 8.41E+02 & 1.35E+03 & 8.14E+02 & 7.58E+02 \\
F24 & 1.21E+03 & 1.14E+03 & 1.09E+03 & 7.24E+02 & 8.16E+02 & 1.30E+03 & 8.78E+02 & 8.30E+02 \\
F25 & 5.12E+02 & 2.72E+03 & 2.22E+03 & 1.06E+03 & 3.08E+03 & 9.11E+02 & 6.34E+02 & 5.65E+02 \\
F26 & 6.09E+03 & 8.76E+03 & 6.33E+03 & 3.95E+03 & 5.82E+03 & 1.05E+04 & 5.47E+03 & 4.60E+03 \\
F27 & 1.07E+03 & 1.12E+03 & 5.00E+02 & 8.99E+02 & 8.75E+02 & 1.69E+03 & 7.22E+02 & 6.95E+02 \\
F28 & 4.97E+02 & 6.77E+03 & 5.00E+02 & 1.48E+03 & 5.11E+03 & 1.20E+03 & 7.09E+02 & 5.87E+02 \\
F29 & 2.55E+03 & 1.04E+04 & 6.25E+03 & 1.62E+03 & 2.31E+03 & 4.85E+03 & 1.67E+03 & 1.31E+03 \\
F30 & 5.58E+07 & 2.43E+09 & 3.38E+08 & 1.09E+08 & 1.29E+08 & 1.35E+08 & 2.31E+06 & 1.51E+06 \\
\hline
\end{tabular}
\end{table*}

\begin{table*}[t!]
\centering
\caption{Experimental results for $D=100$.}
\label{tab7}
\begin{tabular}{c|cccccccc}
\hline
& PSO & CMA-ES & ABC & GWO & MFO & WOA & HMS & MCS-HMS \\ 
\hline 
F1 & 1.07E+09 & 4.20E+10 & 3.61E+11 & 4.25E+10 & 1.24E+11 & 1.98E+07 & 1.51E+10 & 2.89E+07 \\
F2 & 8.45E+71 & 1.04E+166 & 2.96E+182 & 4.30E+133 & 7.99E+159 & 1.15E+148 & 6.53E+134 & 2.10E+109 \\
F3 & 2.36E+05 & 8.50E+05 & 2.22E+06 & 2.45E+05 & 6.55E+05 & 7.55E+05 & 1.14E+05 & 1.51E+05 \\
F4 & 3.87E+02 & 1.71E+04 & 1.53E+05 & 3.36E+03 & 2.38E+04 & 6.69E+02 & 2.01E+03 & 4.34E+02 \\
F5 & 1.09E+03 & 1.07E+03 & 2.03E+03 & 6.41E+02 & 1.20E+03 & 9.25E+02 & 6.78E+02 & 6.48E+02 \\
F6 & 8.60E+01 & 1.44E+01 & 1.39E+02 & 3.65E+01 & 7.32E+01 & 7.93E+01 & 3.32E+01 & 2.24E+01 \\
F7 & 1.25E+03 & 8.65E+02 & 9.19E+03 & 1.25E+03 & 4.04E+03 & 2.57E+03 & 9.67E+02 & 1.13E+03 \\
F8 & 1.20E+03 & 1.27E+03 & 2.08E+03 & 6.33E+02 & 1.27E+03 & 1.12E+03 & 7.14E+02 & 7.20E+02 \\
F9 & 6.32E+04 & 3.25E+04 & 1.93E+05 & 3.27E+04 & 4.21E+04 & 3.54E+04 & 2.22E+04 & 2.05E+04 \\
F10 & 2.20E+04 & 3.04E+04 & 3.26E+04 & 1.57E+04 & 1.68E+04 & 1.95E+04 & 1.79E+04 & 1.60E+04 \\
F11 & 2.44E+03 & 4.51E+05 & 8.88E+05 & 5.33E+04 & 1.32E+05 & 1.17E+04 & 1.35E+04 & 1.08E+04 \\
F12 & 8.27E+08 & 5.27E+10 & 1.19E+11 & 7.40E+09 & 3.48E+10 & 7.23E+08 & 1.52E+09 & 1.71E+08 \\
F13 & 5.08E+07 & 1.16E+10 & 3.13E+09 & 4.15E+08 & 4.43E+09 & 6.20E+04 & 3.78E+07 & 3.73E+04 \\
F14 & 1.57E+06 & 1.15E+08 & 8.61E+07 & 4.95E+06 & 8.14E+06 & 1.45E+06 & 2.07E+06 & 1.58E+06 \\
F15 & 1.52E+07 & 5.73E+09 & 5.70E+08 & 1.02E+08 & 1.81E+09 & 6.21E+04 & 5.00E+05 & 1.24E+04 \\
F16 & 5.85E+03 & 1.20E+04 & 1.52E+04 & 4.42E+03 & 6.36E+03 & 8.43E+03 & 6.06E+03 & 5.17E+03 \\
F17 & 4.32E+03 & 4.22E+04 & 2.97E+04 & 3.54E+03 & 6.44E+03 & 5.47E+03 & 5.25E+03 & 3.98E+03 \\
F18 & 3.18E+06 & 1.39E+08 & 2.86E+08 & 4.83E+06 & 1.63E+07 & 1.88E+06 & 5.67E+06 & 3.21E+06 \\
F19 & 2.67E+07 & 4.44E+09 & 7.29E+06 & 1.15E+08 & 7.69E+08 & 1.33E+07 & 1.54E+06 & 1.44E+04 \\
F20 & 3.77E+03 & 5.14E+03 & 6.57E+03 & 3.28E+03 & 3.78E+03 & 4.23E+03 & 3.65E+03 & 3.42E+03 \\
F21 & 1.69E+03 & 1.44E+03 & 2.56E+03 & 8.69E+02 & 1.55E+03 & 1.79E+03 & 1.10E+03 & 1.05E+03 \\
F22 & 2.44E+04 & 3.11E+04 & 3.34E+04 & 1.74E+04 & 1.84E+04 & 2.17E+04 & 1.99E+04 & 1.73E+04 \\
F23 & 2.91E+03 & 1.84E+03 & 3.16E+03 & 1.23E+03 & 1.49E+03 & 2.44E+03 & 1.41E+03 & 1.12E+03 \\
F24 & 3.04E+03 & 2.44E+03 & 5.16E+03 & 1.75E+03 & 1.94E+03 & 3.55E+03 & 1.95E+03 & 1.67E+03 \\
F25 & 9.45E+02 & 8.27E+03 & 8.55E+04 & 3.42E+03 & 1.12E+04 & 1.13E+03 & 1.59E+03 & 1.01E+03 \\
F26 & 1.71E+04 & 1.93E+04 & 4.96E+04 & 1.22E+04 & 1.53E+04 & 2.82E+04 & 1.58E+04 & 1.28E+04 \\
F27 & 6.43E+02 & 2.00E+03 & 5.00E+02 & 1.33E+03 & 1.28E+03 & 2.27E+03 & 8.75E+02 & 7.95E+02 \\
F28 & 6.38E+02 & 1.91E+04 & 5.00E+02 & 4.73E+03 & 1.62E+04 & 9.13E+02 & 2.37E+03 & 3.18E+03 \\
F29 & 6.94E+03 & 1.34E+04 & 6.10E+04 & 5.28E+03 & 8.22E+03 & 1.11E+04 & 5.04E+03 & 4.30E+03 \\
F30 & 1.07E+08 & 1.02E+10 & 3.27E+09 & 1.03E+09 & 2.33E+09 & 1.83E+08 & 1.02E+07 & 8.14E+04 \\
\hline
\end{tabular}
\end{table*}

Since the employed methods are stochastic, we run all algorithms 25 times and report the mean over these runs in terms of the difference between the optimal value and that returned by an algorithm. The obtained results are given in Tables~\ref{tab5}, \ref{tab6}, and~\ref{tab7} for $D=30$, $D=50$ and $D=100$, respectively, and show impressively that MCS-HMS not only provides superior performance compared to standard HMS but that it also outperforms all the other algorithms.

\begin{figure*}[t!]
\centering
\includegraphics[width=.9\textwidth]{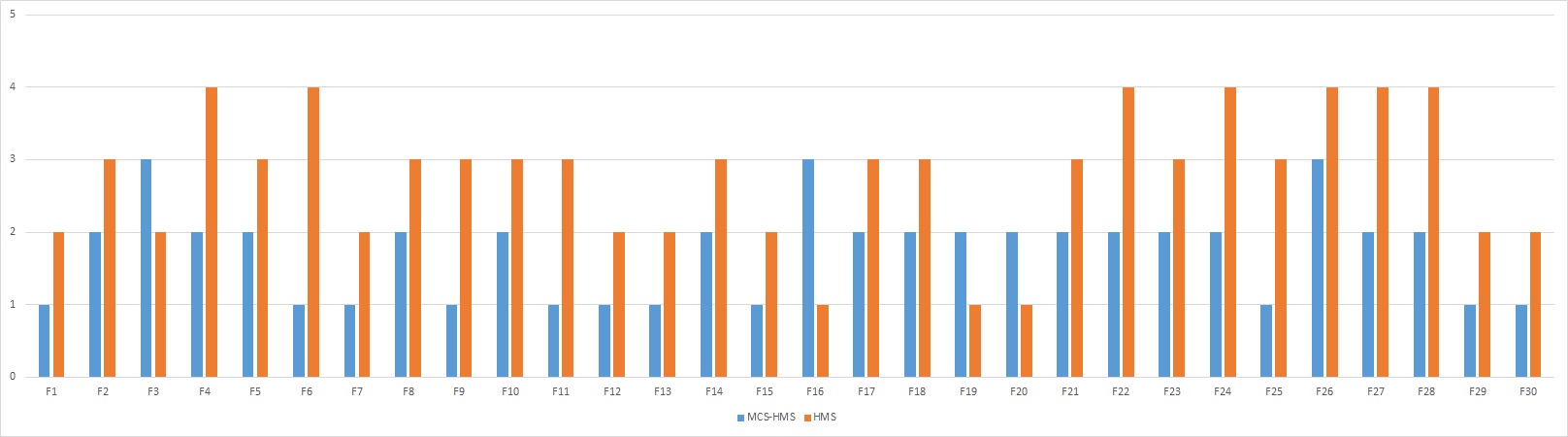}
\caption{Ranking comparison between MCS-HMS and HMS for $D=30$.}
\label{fig2}
\end{figure*}

\begin{figure*}[t!]
\centering
\includegraphics[width=.9\textwidth]{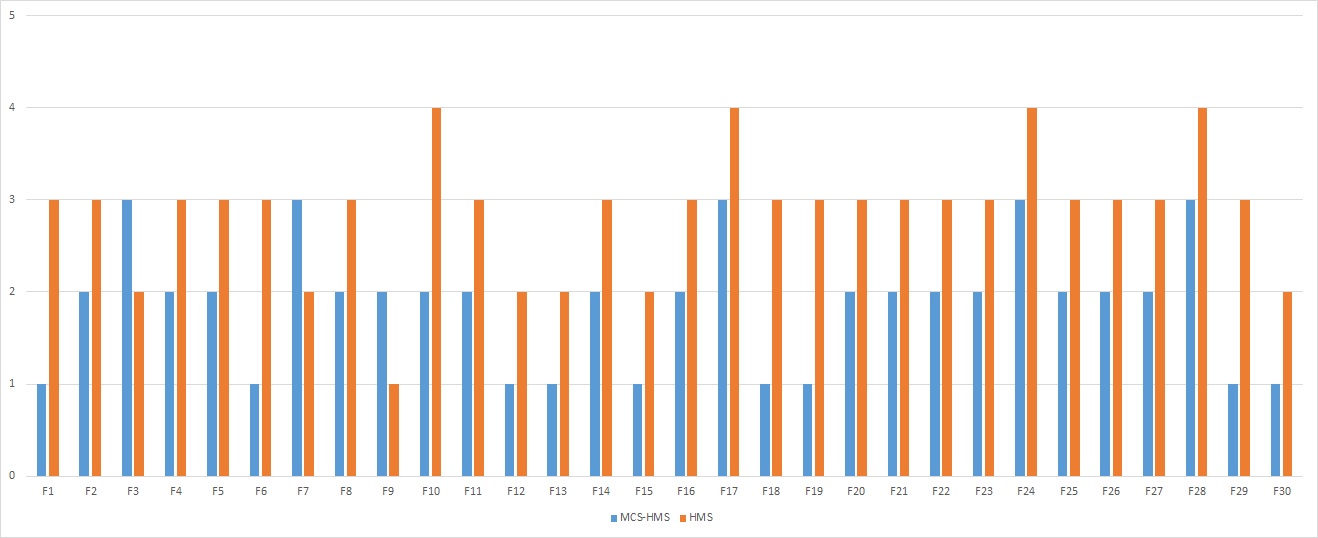}
\caption{Ranking comparison between MCS-HMS and HMS for $D=50$.}
\label{fig3}
\end{figure*}

\begin{figure*}[t!]
\centering
\includegraphics[width=.9\textwidth]{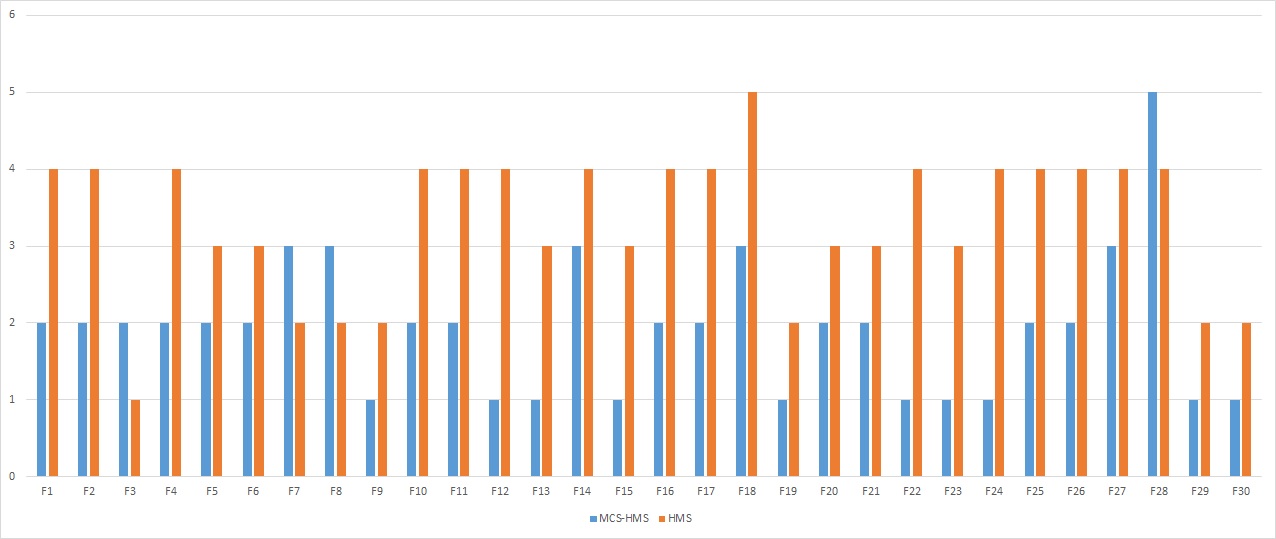}
\caption{Ranking comparison between MCS-HMS and HMS for $D=100$.}
\label{fig4}
\end{figure*}

Its superiority compared to standard HMS is further illustrated in Figs.~\ref{fig2}, \ref{fig3}, and~\ref{fig4} which compared the rankings of HMS and MCS-HMS. From there, we can see that for the vast majority of cases, MCS-HMS is ranked top or second and that in particular, MCS-HMS gives the best result for 24, 25 and 27 functions for $D=30$, $D=50$ and $D=100$, respectively.


\begin{table}[b!]
\centering
\caption{Ranking results for $D=30$.}
\label{tab2}
\begin{tabular}{lcccc}
\hline
& avg.\ rank & best rank & worst rank & std.dev. \\
\hline
PSO     & 3.73         & 1         & 8          & 1.95               \\
CMA-ES  & 7.40         & 4         & 8          & 1.08             \\
ABC     & 5.50         & 1         & 8          & 1.91               \\
GWO     & 3.37         & 1         & 7          & 1.92               \\
MFO     & 5.30         & 3         & 7          & 1.27               \\
WOA     & 6.20         & 3         & 8          & 1.17               \\
HMS     & 2.77         & 1         & 4          & 0.92               \\
MCS-HMS & 1.73         & 1         & 3          & 0.63               \\
\hline
\end{tabular}
\end{table}

Tables~\ref{tab2}, \ref{tab3}, and~\ref{tab4} show, for each algorithm, the average rank, best rank, worst rank and standard deviation over the 30 benchmark functions for $D=30$, $D=50$ and $D=100$, respectively.

\begin{table}[b!]
\centering
\caption{Ranking results for $D=50$.}
\label{tab3}
\begin{tabular}{lcccc}
\hline
& avg.\ rank & best rank & worst rank & std.dev. \\
\hline
PSO     & 3.83         & 1         & 8          & 2.02               \\
CMA-ES  & 6.70         & 1         & 8          & 1.86             \\    
ABC     & 6.23         & 1         & 8          & 1.86               \\
GWO     & 3.27         & 1         & 7          & 1.95               \\
MFO     & 5.30         & 2         & 8          & 1.44               \\
WOA     & 5.93         & 3         & 8          & 1.46               \\
HMS     & 2.87         & 1         & 4          & 0.67               \\
MCS-HMS & 1.87         & 1         & 3          & 0.67               \\
\hline
\end{tabular}
\label{tab3}
\end{table}

\begin{table}[b!]
\centering
\caption{Ranking results for $D=100$.}
\label{tab4}
\begin{tabular}{lcccc}
\hline
& avg.\ rank & best rank & worst rank & std.dev. \\
\hline
PSO     & 3.90         & 1         & 7          & 1.90               \\
CMA-ES  & 6.23         & 1         & 8          & 1.87               \\
ABC     & 7.13         & 1         & 8          & 1.93           \\
GWO     & 3.47         & 1         & 6          & 1.84               \\
MFO     & 5.60         & 3         & 7          & 1.31               \\
WOA     & 4.43         & 1         & 8          & 1.94               \\
HMS     & 3.30         & 1         & 5          & 0.94               \\
MCS-HMS & 1.93         & 1         & 5          & 0.89               \\
\hline
\end{tabular}
\label{tab4}
\end{table}

As can be seen from there, MCS-HMS clearly yields the best performance for all dimensionalities, and by a significant margin. Table~\ref{tab8} further illustrates this by a pairwise comparison of MCS-HMS with all other methods. It is evident from there that in the large majority of cases, MCS-HMS outperforms the other approaches.
 
 \begin{table}[t!]
\centering
\caption{Pairwise comparisons between MCS-HMS and the other algorithms.}
\label{tab8}
\begin{tabular}{l|cc|cc|cc}
\hline 
& \multicolumn{2}{c|}{$D=30$} & \multicolumn{2}{c|}{$D=50$} & \multicolumn{2}{c}{$D=100$} \\
MCS-HMS vs. & better & worse & better & worse & better & worse \\
\hline
PSO & 23 & 7 & 23 & 7 & 22 & 8 \\
CMA-ES & 30 & 0 & 28 & 2 & 28 & 2 \\
ABC & 29 & 1 & 28 & 2 & 28 & 2 \\ 
GWO & 20 & 10 & 19 & 11 & 22 & 8 \\
MFO & 30 & 0 & 29 & 1 & 30 & 0 \\
WOA & 30 & 0 & 30 & 0 & 26 & 4 \\
HMS & 26 & 4 & 27 & 3 & 26 & 4 \\
\hline 
\end{tabular}
\end{table}

Last not least, we conduct a two-sided Wilcoxon signed rank test to see if the differences between MCS-HMS and its opponents are statistically significant and give the results in Table~\ref{tab9}. With all $p$-value below 0.05, it is apparent that MCS-HMS statistically outperforms all other algorithms.

\begin{table}[t!]
\centering
\caption{Results ($p$-values) of two-sided Wilcoxon signed test.}
\label{tab9}
\begin{tabular}{lccc}
\hline
& $D=30$  & $D=50$ & $D=100$ \\
\hline
MCS-HMS vs.\ PSO & 0.0148 & 0.0034 & 0.0132 \\
MCS-HMS vs.\ CMA-ES & 1.73E-06 & 4.73E-06 & 2.35E-06 \\
MCS-HMS vs.\ ABC & 2.13E-06 & 2.88E-06 & 4.29E-06 \\ 
MCS-HMS vs.\ GWO & 0.0082 & 0.0057 & 8.31E-04 \\
MCS-HMS vs.\ MFO & 1.73E-06 & 1.92E-06 & 1.73E-06 \\
MCS-HMS vs.\ WOA & 1.73E-06 & 1.73E-06 & 0.003 \\
MCS-HMS vs.\ HMS & 5.29E-04 & 1.15E-04 & 7.51E-05 \\
\hline
\end{tabular}
\end{table}

\section{Conclusions}
\label{sec:Concl}
In this paper, we have introduced MCS-HMS, an enhanced version of the human mental search (HMS) optimisation algorithm. MCS-HMS employs a multi-cluster strategy in HMS's grouping phase which allows for improved exploration ability, while also using a one-step $k$-means algorithm for clustering to speed up the algorithm's execution. Based on a set of experiments carried out on the CEC 2017 test functions, we show that MCS-HMS not only significantly improves the standard HMS algorithm but that it also outperforms a number of other population-based metaheuristics.

\balance

\bibliographystyle{IEEEtran}
\bibliography{Bibfile.bib}

\end{document}